\documentclass[10pt,twoside]{article}

\usepackage{times}
\usepackage[utf8]{inputenc}
\usepackage[T1]{fontenc}
\usepackage{graphicx}

\usepackage{siunitx}
\usepackage{taln2021}
\usepackage[french]{babel} 

\title{Lemmatization of Historical Old Literary Finnish Texts in Modern Orthography}

\author{Mika Hämäläinen\up{1}\quad Niko Partanen\up{2}\quad Khalid Alnajjar\up{1}\\
 {\small
    (1) Department of Digital Humanities \\ 
    (2) Department of Finnish, Finno-Ugrian and Scandinavian Studies \\ 
     (1,2) University of Helsinki, Finland \\ 
    \texttt{
      firstname.lastname@helsinki.fi 
}}}

\begin{document}
\maketitle

\resume{
Les textes écrits en vieux finnois littéraire représentent la première œuvre littéraire jamais écrite en finnois à partir du XVIe siècle. Il y a eu plusieurs projets en Finlande qui ont numérisé des anciennes collections de textes et qui les ont rendues disponibles pour la recherche. Cependant, l'utilisation de méthodes TAL modernes avec telles données pose de grands défis. Dans cet article, nous proposons une approche pour normaliser et lemmatiser simultanément des textes écrits en vieux finnois littéraire à l'orthographe moderne. Notre meilleur modèle donne une précision de 96,3 \% avec les textes écrits par Agricola et de 87,7 \% avec d'autres textes contemporains hors du domaine. Notre méthode est publiée gratuitement sur Zenodo et Github.
}

\abstract{}{
  Texts written in Old Literary Finnish represent the first literary work ever written in Finnish starting from the 16th century. There have been several projects in Finland that have digitized old publications and made them available for research use. However, using modern NLP methods in such data poses great challenges. In this paper we propose an approach for simultaneously normalizing and lemmatizing Old Literary Finnish into modern spelling. Our best model reaches to 96.3\% accuracy in texts written by Agricola and 87.7\% accuracy in other contemporary out-of-domain text. Our method has been made freely available on Zenodo and Github.
}

\motsClefs
  {données historiques, normalisation, lemmatisation}
  {historical data, normalization, lemmatization}

\section{Introduction}

Finnish language has a long literary history starting from the 16th history. 
A large portion of the books printed in Finnish is currenlty openly available for research, and especially the bibliographic record they form has already been studied. 
\cite{lahti2019bibliographic} investigated the document size and language in the bibliographic metadata records, and  \cite[58]{tolonen2019quantitative} took into account a number of other metadata fields including the titles, and recognize the lack of full-text documents as a downside of investigation that is possible. 
Although the whole body of books printed in Finland is not available as full-text, there are numerous smaller corpora that can already be used in various ways. 

Both the National Library of Finland and Institute for the Languages of Finland have produced a large number of digitized material from the era of the Old Literary Finnish. 
Former has made a large amount of scanned and text recognized publications available \cite{digi}, and latter has created large plain text corpora from selected works \cite{vks}. 
The Agricola corpus used in this study has been morpho-syntactically annotated, and is to our knowledge the only annotated resource in the Old Literary Finnish  \cite{agricola-v1-1-korp_fi}. 
Another very central resources is the Dictionary of Old Literary Finnish.\footnote{https://kaino.kotus.fi/vks/} 
Thus far the Old Literary Finnish has largely eluded any attempts of NLP research as the historical written form cannot be processed easily with currently available NLP tools for Finnish, as they are designed for modern Finnish. We present our approach for normalizing and lemmatizing Old Literary Finnish automatically to modern Finnish orthography. As the resources for processing historical Finnish text are scarce, we have released the models presented in this paper on Zenodo\footnote{https://zenodo.org/record/4734143} and through an easy-to-use Python library\footnote{https://github.com/mikahama/murre}.

The use of existing NLP tools targeted for modern Finnish on historical materials can be made possible through normalization. Previous work conducted on English data indicates that normalization is a viable way of improving the accuracy of NLP methods such as POS tagging \cite{van2017normalize}. Another direction of digital humanities study has benefited from normalization of historical data in studying the use of neologisms in old letters \cite{14fd5fbd7da8429bb72e9e67e6b81535,tanja-multlingual}. In their approach, without normalization, they would have been able to cover only a small subset of the corpus. The same corpus has also been studied without NLP tools \cite{terttu-multlingual}.

The history of printed written Finnish starts from the 16th century with the works of Mikael Agricola. 
His primer and religious works were followed by a continuous increase in the amount of the written Finnish materials. 
The language form used by Agricola is known as Old Literary Finnish (\textit{vanha kirjasuomi}). 
The majority of the early Finnish publications included religious materials, although the text types started to diversify already in the 18th century. 
The majority of the oldest texts are translations. 
The period of Old Literary Finnish is often estimated to have lasted until 1810, after which the written Finnish started to transform into Early Modern Finnish. 
Eventual changes in printing laws and printing technology, which expanded the amount and variation of the printed materials, and the creation of regular Finnish newspapers, contributed to the stabilization of the written standard. 

Linguistically one of the exceptional features of Old Literary Finnish is the variation it displays. 
The orthography was not yet entirely established, and there was extensive spelling variation.  
The age of these materials adds also a historical dimension, as there are linguistic features that are not present in the modern Finnish, or exist currently only in the dialects. 

Our main contributions in this work are:
\begin{itemize}
    \item Building the first artificial neural network model for normalizing historical Finnish.
    \item Conducting an evaluation for assessing the performance of the model on historical data from 1) the same source of the data used in building the model and 2) external out-of-domain historical data. In both cases, a high accuracy is achieved by the model.
    \item Publishing a user-friendly Python library that permits an instant usability of our model.
\end{itemize}

The target language form in our work is modern Finnish. 
Our model primarily lemmatizes, but since the output is harmonized into contemporary orthographic forms, the work is closely connected to the normalization task as well, and the output can be considered as one type of a normalization. 
The exact lemmatization choices and conventions were decided at the level of the original morpho-syntactic database, and we followed those closely also when our own additional test material was created. 
In later research also the morpho-syntactic annotations present in the corpus could be taken into account to further enrich the analysis. 
However, at the current stage a successful lemmatization is already a large improvement in available NLP methods. 

This paper is structured as follows. We begin by describing the related work. Thereafter, the details of the data used to build the neural model are given, followed by the architecture and hyperparameters of the neural model. Section 5 presents the results and evaluation where we explain the different training strategies we experimented with and their performance against a baseline (historical Omorfi). Lastly, we discuss and conclude our work while highlighting future directions with a potentially great impact on humanities research such as automatic analysis of historical Finnish.

\section{Related Work}

Historical text normalization has been studied in the past for other languages than Finnish. A recent literature review \cite{bollmann-2019-large} finds that there are five categories in which modern normalization approaches can be divided: substitution lists like VARD \cite{rayson2005vard} and Norma \cite{bollmann2012-semi}, rule-based methods \cite{baron2008vard2,porta2013edit}, edit distance based approaches \cite{hauser2007unsupervised,amoia2013using}, statistical methods and most recently neural methods \cite{partanen-et-al,duong2020unsupervised}.

Statistical machine translation (SMT) based methods have been the most successful ones in the past in terms of statistical methods. The key idea behind these methods is to approach the task as a character-level machine translation problem, where a word is translated character by character to its normalized form. These methods have been applied to historical text \cite{pettersson2013smt,hamalainen2018normalizing} and dialect normalization \cite{swissgerman}.

In the recent years, normalization has been approached as a character-level neural machine translation (NMT) problem similarly to the previous SMT approaches. The additional advantage is that a neural model does not need a separate language model like SMT does. \citet{bollmann-sogaard-2016-improving} have shown that a bi-directional long short-term memory (bi-LSTM) can be used to normalize historical German texts. The paper presents a so-called multi-task learning setting where auxiliary data is added to improve the performance of the model. Multi-task learning setting generally improved the results. Their system outperformed the existing conditional random fields and Norma based approaches in terms of accuracy. 

Text written in Uyghur language has been normalized with an LSTM and a noisy channel model (NCM) \cite{tursun-cakici-2017-noisy}. They use a relatively small set of gold annotated data for training (around 200 hand normalized social media sentences). They augment this data by synthetically generating non-normalized text by introducing random changes in normalized text.  In the same fashion, another research has used an LSTM model to normalize code-mixed data \cite{mandal-nanmaran-2018-normalization}.

Recently \citet{hamalainen-etal-2019-revisiting} have shown that bi-directional recurrent neural networks (BRNN) outperform regular unidirectional recurrent neural networks (RNN) when normalizing historical English data. Interestingly, additional layers and different attention models do not improve the results. Additional data such as time period, social metadata or pronunciation information in IPA characters makes the results worse. According to them, post-processing can boost the accuracy of a character level NMT model more than changing the network structure. A simple dictionary filtering method improved the results.

Omorfi \cite{pirinen2015development} is a popular rule-based tool used to do morphological analysis and lemmatization of modern Finnish. While Omorfi itself is not relevant for our work, there is a GitHub fork of the project known as Historical Omorfi\footnote{\url{https://github.com/jiemakel/omorfi/}}. The fork introduces several improvements to better cater for historical Finnish text. Currently, this tool is the only tool available for lemmatizing historical Finnish. Thereby we compare the results of our model to this tool.

\section{Dataset of Historical Finnish}

In order to use machine learning methods, data is needed to train a model. The data needs to have text written in Old Literary Finnish and its normalized lemmas. The lemmas should be aligned on a word level with the historical data in order to train the normalization more accurately. 
Fortunately, such a dataset exists. The corpus we use is \textit{The Morpho-Syntactic Database of Mikael Agricola's Works} \cite{agricola-v1-1-korp_fi} that contains 522,237 tokens and 38,222 sentences. 
The corpus includes all nine Finnish books translated by Mikael Agricola. 
The corpus is openly available in the Language Bank of Finland.\footnote{\url{urn:nbn:fi:lb-2019121804}} 
In our testing we also use the Dictionary of Old Literary Finnish\footnote{\url{https://kaino.kotus.fi/vks}} \cite{vks}, from which a small number of sentences has been sampled and manually normalized and lemmatized. 
Both resources are available under Creative Commons licenses. 
Since the \textit{The Morpho-Syntactic Database of Mikael Agricola's Works} is licensed under CC BY-ND 4.0 (CLARIN PUB), we do not redistribute the training material ourselves, but it can be accessed in the Language Bank of Finland's concordance service\footnote{\url{https://korp.csc.fi}}. 
Naturally, since this data is so old, it is already in Public Domain, and many of the original works are entirely openly accessible. 
For example, the Agricola's prayer book is available as high quality scans by the National Library of Finland\footnote{\url{https://www.doria.fi/handle/10024/43445}}.

Although there has not been prior use of this dataset in the computational linguistics, the corpus has been used in the linguistic studies. 
To illustrate this with few recent examples, \citet{toropainen2018yhdyssanat} investigated the compounds in this variety, and \citet{toropainen2015adjektiivialkuiset} studied nouns that contain an initial adjective. 
\citet{salmi2020german} discussed recently the German influence in the Agricola's language. 
Also annotating the corpus into it's current stage has been a long undertaking, and \citet{inaba2015suomen} investigated the use of two Finnish cases with a prototype of the current database. 
It is beyond doubt that the materials of Old Literary Finnish still have much to contribute to the linguistic research. 
We believe that by creating new tools for natural language processing of these and similar materials we can further expand toward these goals. 
The research concerning Agricola's language and Old Literary Finnish in general is naturally much wider and has a long research history, especially in Finland, and we primarily wanted to illustrate in this section some of the previous studies where the same database was used. 

\section{Neural Normalization}
In this section, we describe our artificial neural network model for normalizing and lemmatizing historical Finnish into standard Finnish. We begin by explaining how the dataset is used and how it is preprocessed. Thereafter, we elucidate the architecture of the model along with the technical details of its hyperparameters.

The corpus contains nine distinct works. 
In order to test the model's accuracy realistically, we selected seven books into the training data, and left the remaining two into the test set. 
Thereby the training data was 393,779 tokens and the test set 128,294 tokens. 
The books in the test set were specifically `Messu eli Herran echtolinen' from 1549 and `Rucouskiria' from 1544. 
Additionally 15\% of the training data was used in the validation set. 
We considered it important not to select the test set from the entire corpus, as this would not give a clear picture about how much the model generalizes into new use cases, which would be the other books written in the same language variety. 
With the current sparsity of manually annotated data, the Agricola works in the currently used corpus, but which we kept unseen for the model, were the best option we had. 
We also extended the testing into other examples of Old Literary Finnish, but in a smaller scale. 
The data has the original written form of each sentence and a token level normalized lemma for each word. We use this parallel data to train our models. 

We model the problem as a character level NMT problem. In practice, we split words into characters separated by white space and mark actual spaces between words with an under score (\_). 
This allows to pass word boundary information to the model while the characters themselves are separated by spaces. 
We train the model to predict from the Old Literary Finnish word forms to the lemmas. As previous research \cite{partanen-et-al} has found that using chunks of words instead of full sentences at a time improves the results, we train different models with different chunk sizes. 
This allows direct comparison of different chunk sizes. 
This way we train the models to predict one word at a time, two words at a time all the way to five words at a time. An example of the data can be seen in Table \ref{tab:training}. 
This example is within the sentence A-II-369 in the corpus, which is located in the New Testament translation. 

\begin{table}[h]
\centering
\begin{tabular}{|l|c|c|}
\hline
\textbf{Size}    & \textbf{Source}                                        & \textbf{Target}                                      \\ \hline
\textbf{Chunk 1} & s y d h e m e n                               & s y d ä n                                   \\ \hline
\textbf{Chunk 3} & p a l u e l l e n \_ h e r r a \_ c a i k e n & p a l v e l l a \_ h e r r a \_ k a i k k i \\ \hline
\end{tabular}
\caption{An example of the training data for chunk sizes 1 and 3.}
\label{tab:training}
\end{table}

We train all models using a bi-directional long short-term memory (LSTM) based model \cite{hochreiter1997long} by using OpenNMT-py \cite{opennmt} with the default settings except for the encoder where we use a BRNN (bi-directional recurrent neural network) \cite{schuster1997bidirectional} instead of the default RNN (recurrent neural network), since BRNN based models have been shown to provide better results in a variety of tasks. 
We use the default of two layers for both the encoder and the decoder and the default attention model, which is the general global attention presented by  \citealt{luong2015effective}. The models are trained for the default of 100,000 steps. All models are trained with the same random seed (3435) to ensure reproducibility and to make their intercomparison possible.

\section{Results and Evaluation}

Our initial evaluation results were very good as seen in Table \ref{tab:results} where the accuracies are reported on a word level. The best model was the chunk of 3. 
The quality we reach is very high and on par with other comparable normalization tasks, such as in \cite{partanen-et-al}, which also corroborate the best performance at the chunk of three tokens. 
However, we are also interested in seeing how well our model works with other Old Literary Finnish texts. At any rate, we can see that our models outperform Historical Omorfi, which is the only tool publicly available for historical Finnish. As Omorfi produces all the possible lemmas for a given word, we count the accuracy based on if the correct lemma is in the list of the lemmas Omorfi produced for each word.

\begin{table}[h]
\centering
\begin{tabular}{|l|c|c|c|c|c|l|}
\hline
         & \textbf{Chunk 1} & \textbf{Chunk 2} & \textbf{Chunk 3} & \textbf{Chunk 4} & \textbf{Chunk 5} & \textbf{Omorfi} \\ \hline
\textbf{Accuracy} & 96.1\%  & 96.2\%  & 96.3\%  & 96.2\%  & 95.9\%  & 40.5\% \\ \hline
\end{tabular}
\caption{Token level accuracy of each model in the test data.}
\label{tab:results}
\end{table}

Because there is no other dataset freely available that would both be written in Old Literary Finnish and lemmatized to modern orthography, we take randomly 50 sentences from the example sentences of the dictionary of Old Literary Finnish\footnote{https://kaino.kotus.fi/vks} that is available online. 
Altogether these sentences have 562 words (excluding punctuation). 
We lemmatize these sentences with the model that has been trained with chunks of 3 as it worked the best out of the models and verify the lemmatization by hand. The results of this experiment are seen in Table \ref{tab:res2}.

\begin{table}[h]
\centering
\begin{tabular}{|c|c|c|}
\hline
         & \textbf{Chunk 3} & \textbf{Omorfi} \\ \hline
\textbf{Accuracy} & 87.7\%  & 47.9\% \\ \hline
\end{tabular}
\caption{Accuracy in the Old Literary Finnish dictionary sample.}
\label{tab:res2}
\end{table}

The results drop in this evaluation, but it is only to be expected given that out-of-domain performance is typically lower for neural models. Nevertheless, we see clearly that the model does well in out-of-domain data and beats the current state of the art. The fact that Historical Omorfi gets better results in this dataset is a good indication that the text is very different from what is in the Agricola dataset. 
However, it clearly is not entirely distinct when we consider how well the model still performed. 

If we look at the results more closely, analyzing the errors, we can see that the model usually does not predict non-words but rather words that are a part of the Finnish vocabulary. This indicates that the model has learned a good target representation. There are several errors in which the model has normalized the historical word correctly, but it has not lemmatized it, for example \textit{runsast} was normalized to \textit{runsaasti} although the lemma would be \textit{runsas} `plenty'. Another example is \textit{ulosteon} that was already written as in the modern orthography was normalized unchanged to \textit{ulosteon}, while the lemma is \textit{ulosteko}. 
This example also illustrates how there is variation and historical change in spelling, as we find in the Agricola corpus comparable words spelled with the initial \textit{v}, whereas in the later materials we find the variant above that is spelled closely the current standard. 
Analysing how this relates to the changes in characters used in different centuries is beyond the scope of our study, but it illustrates well the kind of variation we can find in the historical texts and their digitized versions. 

To analyze the errors further, the most typical source of problems are verbs that get lemmatized into nouns that look similar and vice versa. Thereby \textit{olan} was lemmatized as \textit{olla} `to be' while it should have been \textit{olka} `shoulder'. Also, \textit{nuole} was lemmatized as \textit{nuoli} `arrow', while the correct lemma is \textit{nuolla} `to lick'.  
Normalization to a wrong lemma within the same part-of-speech is also possible e.g. \textit{kaipanne} to \textit{kaivaa} `to dig' instead of \textit{kaivata} `to miss'. 
Improving the recognition of such instances is a very important task for the future work, but the current accuracy also appears to be useful and satisfactory for many tasks, and is without doubt an improvement to the existing methods.

\section{Conclusions and Future Work}

Our results have a clear indication, both with in-domain and out-of-domain test data, of working successfully in lemmatizing Old Literary Finnish in the modern orthography. The models have been released on Zenodo and in a Python library\footnote{https://github.com/mikahama/murre}. By sharing the models we are making NLP research on historical Finnish data more widely accessible for the research community, as the currently available Historical Omorfi does not work well for texts that are this old. 
Our study also creates a benchmark into which the further work can easily be compared. 

Having lower accuracy in another dataset is a reminder of the importance of evaluating normalization models on data that comes from a different distribution. This is something seldom seen in the previous work on historical spelling normalization. Despite this, the accuracy has remained relatively high and we have identified several possibly problematic phenomena in the test data that are more prone to errors. This error analysis helps in understanding the biases that using our model might introduce in historical data when it is used to lemmatize a corpus completely new to the model. 
Further research is needed to evaluate how the error rate varies when the distance grows to the materials of Agricola. 
It is beyond doubt that more diverse training material is needed to successfully process the entire corpus of Old Literary Finnish, but our study certainly has improved the position to initiate and continue such work. 

The work we presented here also makes it possible for the current research on analyzing historical Finnish newspapers, such as~\cite{10.1145/3383583.3398627,OCRhistorical}, to standardize post-OCR historical Finnish, which in turn permits employing state-of-the-art Finnish NLP methods and tools on such data (e.g. sentiment and semantic analysis~\cite{hamalainen-alnajjar-2019-lets}).

When we work with the currently available resources, we must also remain aware that the digital versions have been edited in various ways, and do not contain necessarily all features of the original printed text \cite[175]{toropainen2016typografian}. 
Now when we increasingly have access also to the original prints as high quality scans, it is important to think how these different resources can be connected to one another. 
This will need a combination of both text recognition and NLP tools. 

In the future, we are interested in conducting work on semantic change on historical data. This should be greatly facilitated by the fact that we can now considerably reliably lemmatize historical text. This means that training word embeddings models will become more accurate as the model is trained on lemmas instead of inflectional forms.

\bibliographystyle{taln2021}
\bibliography{biblio}

\end{document}